\pdfoutput=1

\documentclass[11pt]{article}

\usepackage[]{EACL2023}

\usepackage{times}
\usepackage{latexsym}

\usepackage[T1]{fontenc}

\usepackage[utf8]{inputenc}
\usepackage{graphicx}
\usepackage{graphicx,xcolor}
\usepackage{multirow}
\usepackage{caption}
\usepackage{subcaption}
\usepackage{amsfonts}
\usepackage{amsmath}
 \usepackage{mathtools}  
\usepackage{subcaption}
\usepackage{xcolor, colortbl}
\usepackage{booktabs}
\usepackage{algorithm}
\usepackage{algorithmic}
\usepackage[utf8]{inputenc}
\usepackage{epsfig}
\usepackage{graphicx}
\usepackage{amssymb}
\usepackage{epsfig}
\usepackage{amsmath}
\usepackage{multirow}

\usepackage{microtype}

\usepackage{inconsolata}

\title{\textit{READ}: Reinforcement-based Adversarial Learning for Text Classification with Limited Labeled Data}

\author{ Rohit Sharma \quad Shanu Kumar \quad Avinash Kumar\\
Microsoft Corporation, India \\
{\tt \small rohit.sharma@alumni.iitgn.ac.in} {\tt \small \{shankum,avkum\}@microsoft.com} }

\begin{document}
\maketitle
\begin{abstract}
Pre-trained transformer models such as BERT have shown massive gains across many text classification tasks. However, these models usually need enormous labeled data to achieve impressive performances. Obtaining labeled data is often expensive and time-consuming, whereas collecting unlabeled data using some heuristics is relatively much cheaper for any task. Therefore, this paper proposes a method that encapsulates reinforcement learning-based text generation and semi-supervised adversarial learning approaches in a novel way to improve the model's performance. Our method \textit{READ} (\textbf{RE}inforcement-based \textbf{AD}versarial learning) utilizes an unlabeled dataset to generate diverse synthetic text through reinforcement learning, improving the model's generalization capability using adversarial learning. Our experimental results show that \textit{READ} outperforms the existing state-of-art methods on multiple datasets. 

\end{abstract}

\section{Introduction}
\label{submission}

The introduction of pre-trained transformer-based large-scale models such as BERT \cite{devlin-etal-2019-bert}, GPT-2 \cite{radford2019language}, and RoBERTa \cite{liu2019roberta} has led to impressive results on many Natural Language Processing (NLP) tasks. However, even with a pretraining these models require a large number of labelled data for fine-tuning on a downstream task \cite{yogatama2019learning, croce2020gan}. Few works \cite{mukherjee2020uncertainty, croce2020gan} have shown significant drop in performance while fine-tuning BERT using only limited examples.

Obtaining labelled data can be expensive and time-consuming process \cite{dandapat-etal-2009-complex, sabou2012crowdsourcing, fort2016collaborative}, nevertheless collecting unlabeled data for any downstream task is relatively much cheaper. Semi-supervised learning \cite{kipfsemi, zhu2005semi} has been shown to be one of the promising paradigms to generalize even with few labelled data, by utilizing large amounts of unlabeled data. Recently, \citet{miyato2016adversarial, xie2020unsupervised, izmailov2020semi, liu2021flitext} have shown substantial improvements for text classification tasks using consistency training on unlabeled data via data augmentations such as back-translation. One of these approaches is Semi-Supervised Generative Adversarial Networks (SS-GANs) \cite{salimans2016improved}, which uses GANs \cite{goodfellow2014generative} to expose the huge amounts of unlabeled to the classifier for improving generalization capability. GAN-BERT \cite{croce2020gan} extends SS-GANs by training BERT with unlabeled data in a generative adversarial setting and achieves comparable results even with less than 200 labeled examples to a fully supervised setting.

GAN-BERT employs a generator which produces features resembling the real data distribution due to adversarial training, while a discriminator is trained to assign class categories and to distinguish samples of the generator from the real instances. The adversarial training helps GAN-BERT to learn generalizable feature representations. We hypothesize that adversarial learning with synthetic feature representations may not fully unlock generalization capabilities of pre-trained models and argue that generating text instead of feature representations can further improve their generalization capabilities. The feature generator in GAN-BERT is only used during training and becomes unusable during inference, whereas text generators can help in debugging and model explainability. 


In last decade, various methods \cite{wiseman2016sequence, dong2019unified, song2019mass, lewis2020bart} have been proposed for text generation, however in this work, we employ inverse reinforcement learning (IRL) \cite{ijcai2018-0606} framework for text generation which alleviate the problem of mode collapse and reward sparsity. IRL generates text using a reward function which gives a higher reward to the real texts and lower rewards to the generated texts. We could have used the text generated from IRL along with the unlabeled data in GAN-BERT, instead we propose \textit{READ} which bridges both adversarial training and text generation through IRL for improving the generation capability of pre-trained models. 


The contributions of this paper can be summarized as follows: 
1) We propose \textit{READ} to fully unlock the generalization capability of pre-trained models through reinforcement-based text generation with adversarial learning.
2) Through experiments, we show that our method outperforms existing semi-supervised methods on multiple datasets using various pre-trained models.
3) We perform extensive analysis to show the importance of encapsulation of text generation and adversarial learning.
4) We empirically demonstrate the generalization capability and text generation quality of \textit{READ}.

\section{Methodology: \textit{READ}}
Assuming we have a small labeled dataset $L$ and an unlabeled dataset $U$, the aim is to train a classifier over $k$-class objective using a pre-trained model on the dataset $L$. Similar to GAN-BERT, we propose \textbf{\textit{READ}} to improve the performance it using the unlabeled set $U$. \textit{READ} consists of following components: Text Generator $\mathcal{G}$, Reward Approximator $\mathcal{R}$, pre-trained Transformer Model $\mathcal{M}$ and Classifier $\mathcal{C}$. In next subsections, we will explain these components and their objectives.



\begin{figure}
  \centering
  \includegraphics[scale=0.065]{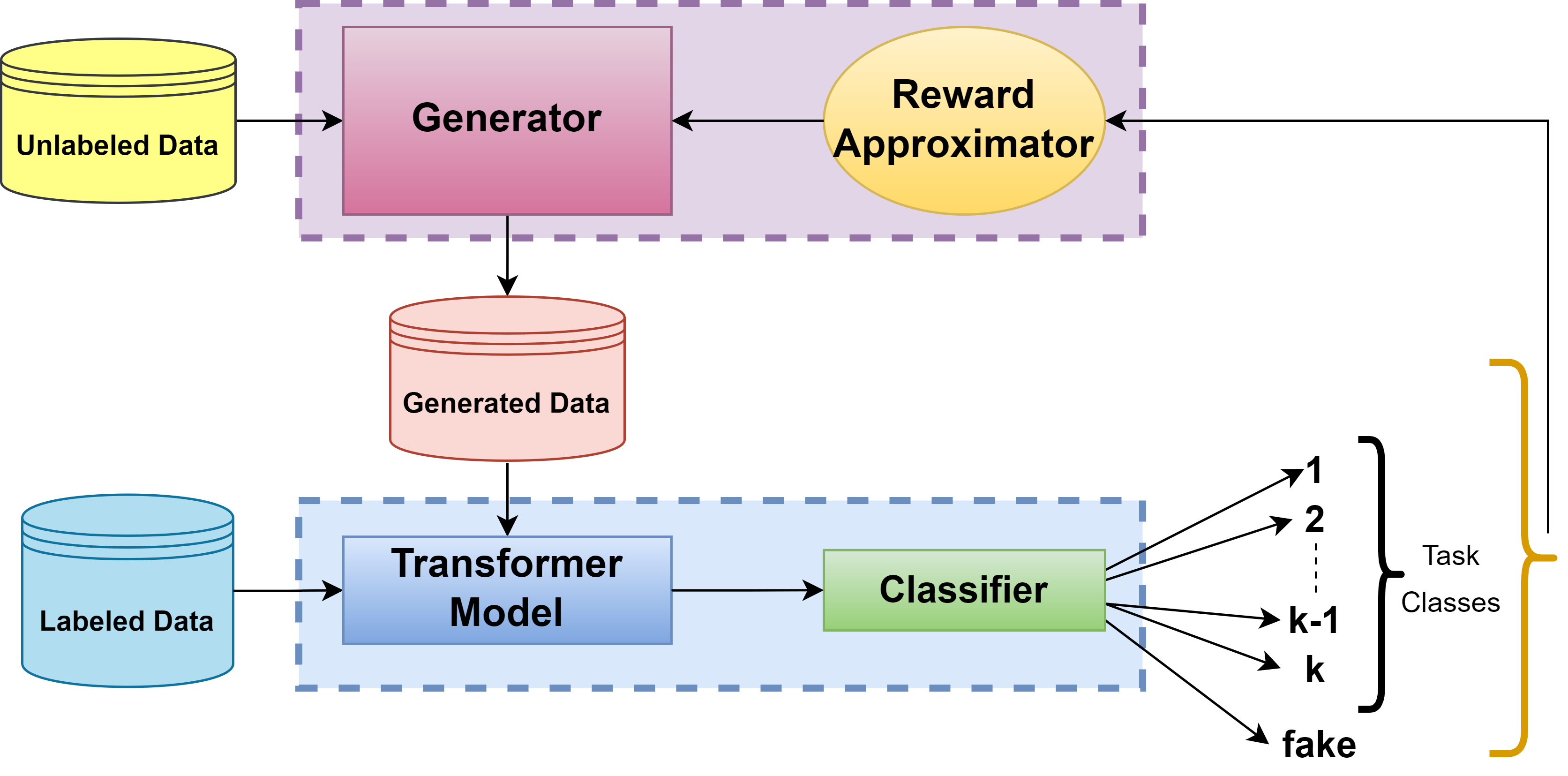}
  \vspace{-1.5em}
  \caption{Architecture of \textit{READ}}
  \label{fig:Plot_RL-BERT_Architecture}
\vspace{-0.5em}
\end{figure}

\textbf{Text Generator $\mathcal{G}$} is implemented by following the IRL, where it trained using the unlabelled dataset $U$ to generate synthetic and diversified texts $U^{\prime}$. Text generation task can be regarded as the generation of the text sequence $x_{1: T}=$ $x_{1}, x_{2}, \cdots, x_{T}$ with a trajectory $\tau = \{s_1, a_1, s_2, a_2, \cdots, s_T, a_T\}$, where $s_t$ is the current state of the previous prediction $x_{1:t}$ and $a_t$ is the action to select the next word $x_{t+1}$. $\mathcal{G}$ is trained to generate real-like examples by maximizing the expected reward $\mathcal{R}(\tau)$.

\textbf{Reward Approximator $\mathcal{R}$} is also defined following IRL as the summation of the rewards of each step with a modification. The reward function at step $t$ is defined using $r_\phi(s_t, a_t)$, where $r_\phi$ is a feed-forward neural network. To bridge the generation and classification processes, we use the  probability $p_{k+1}$ of being classified as fake by the classifier $\mathcal{C}$ as the additional input in the reward function $r_{\phi}\left(s_{t}, a_{t}, p_{k+1}\right)$. The overall reward for a trajectory $\tau$ can be defined as follows:
\begin{equation*} \label{eq:rewardapprox}
R_{\phi}(\tau)=\sum_{t} r_{\phi}\left(s_{t}, a_{t}, p_{k+1}\right)
\end{equation*}

$\mathcal{R}$ is trained to maximize the log-likelihood of the samples in the $U$, whereas $\mathcal{G}$ is trained to maximize the expected reward with an entropy regularization term. We follow IRL framework for defining the training objectives of $\mathcal{R}$ and $\mathcal{G}$, and refer the readers to their work for additional details.

\textbf{Transformer Model  $\mathcal{M}$} is a pre-trained transformer model to encode any input text to a $d$-dimensional feature representation $h \in \mathbb{R}^d$. \textbf{Classifier $\mathcal{C}$} is defined by following GAN-BERT, where $\mathcal{C}$ is trained to classify any feature representation $h$ in one of the $k$ task categories or into the ${k+1}^{th}$ \textit{fake} category, if the $h$ corresponds to a fake example. 

\textbf{Training objective of $\mathcal{M}$ and $\mathcal{C}$} is defined by minimizing following three losses:  $\mathcal{L}_{l}$ loss on classifying the samples from the labeled dataset $L$ into one of the $k$ classes,  $\mathcal{L}_{r}$ loss for not classifying the samples from $L$ and $U$ as \textit{fake}, and an additional $\mathcal{L}_f$ loss for classifying the generated samples from $U^{\prime}$ as \textit{fake}.
\vspace{-0.5em}
\begin{gather*}
\mathcal{L}_{l} = -\mathbb{E}_{x, y \sim L} \log \left[p(\hat{y}=y \mid x, y \in(1, \ldots, k))\right] \\
\mathcal{L}_{u} = -\mathbb{E}_{x \sim L \cup U} \log \left[1 - p(\hat{y}=y \mid x, y = k +1)\right] \\
\mathcal{L}_{f} = -\mathbb{E}_{x \sim U^{\prime}} \log \left[p(\hat{y}=y \mid x, y = k +1)\right]
\end{gather*}
 where $p$ is the probability vector returned by $\mathcal{C}$ for the input $x$.

\begin{figure*}[!ht]
     \centering
     \begin{subfigure}[b]{0.32\textwidth}
         \centering
         \includegraphics[width=\textwidth]{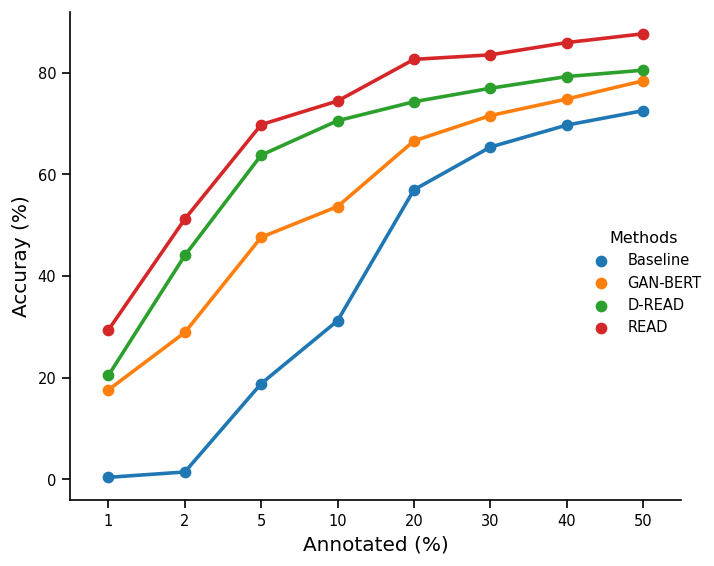}
         \caption{TREC-CF}
         \label{fig:trec_cf_bert}
     \end{subfigure}
     \hfill
     \begin{subfigure}[b]{0.32\textwidth}
         \centering
         \includegraphics[width=\textwidth]{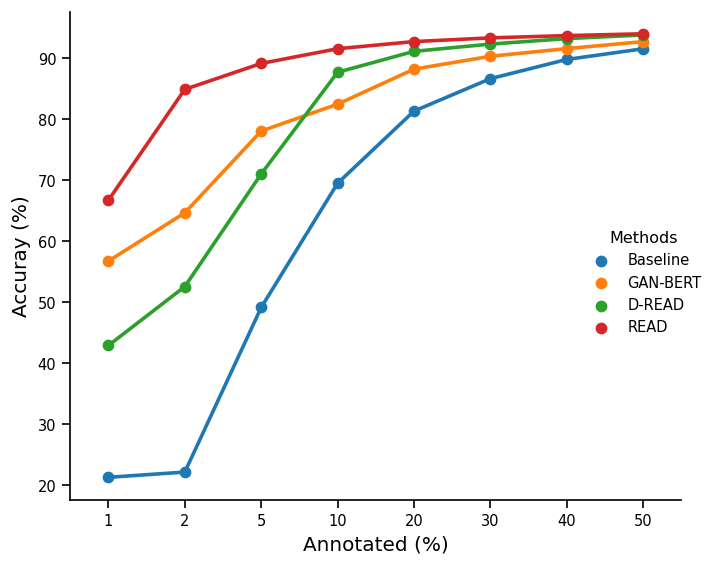}
         \caption{TREC-CC}
         \label{fig:trec_cc_bert}
     \end{subfigure}
     \hfill
     \begin{subfigure}[b]{0.32\textwidth}
         \centering
         \includegraphics[width=\textwidth]{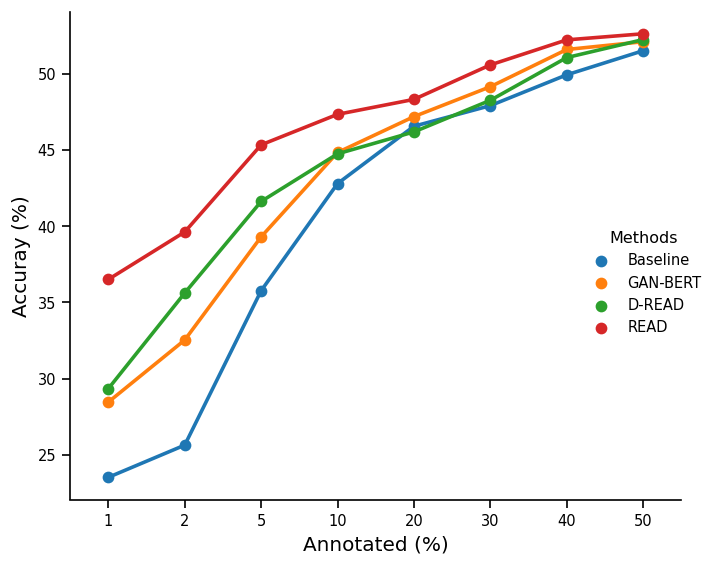}
         \caption{SST-5}
         \label{fig:sst5_bert}
     \end{subfigure}
     \vspace{-0.5em}
      \caption{Accuracy comparison using BERT model.}
        \label{fig:bert_results}
\end{figure*}

In IRL, the reward function is defined using only current and previous states, whereas in \textit{READ}, the reward function also takes the probability of being fake $p_{k+1}$. As, we are training  $\mathcal{M}$ and $\mathcal{C}$ for identifying the generated samples as the fake class, the probability $p_{k+1}$ of being fake will be high for generated examples whereas it will be low for real text samples. Due to this property, reward function in \textit{\textbf{READ}} will encapsulate the classifier's knowledge along with the real text distribution. Simultaneously, we are training the text generator $G$ to maximize the expected reward using adversarial learning to encourage the generation of samples that are not only  similar in form of states but also having low probability of being identified as fake by $\mathcal{C}$ by fooling the classifier. We hypothesize that this form of adversarial learning based on reinforcement-based text generation instead of feature representation will be more robust towards improving the classifier's generalization capability.

\section{Experimental Details}
\subsection{Dataset}
We have evaluated our method's performance on three sentence classification tasks: Fine Grained Question Classification \textit{\textbf{TREC-QCF}} task \cite{lang1995newsweeder}, Coarse Grained Question Classification \textit{\textbf{TREC-QCC}} task \cite{li2006learning}, and Sentiment Analysis \textbf{\textit{SST-5}} task \cite{socher2013recursive}. We have reported the training and test data statistics in Table \ref{tbl:data} in the Appendix.


\subsection{Baselines and \textit{\textbf{READ}}'s Variants}
In our experiments, we compare \textit{READ} with \textbf{GAN-BERT} and \textbf{Baseline} which is a vanilla fine-tuning method without any adversarial training. We experiment with two pre-trained transformer models BERT \cite{devlin-etal-2019-bert} and RoBERTa \cite{liu2019roberta}. To understand the importance of the encapsulation of text generation and adversarial learning, we experiment with disjoint training of text generation and classifier by removing the probability of being fake from the reward function, $r_{\phi}(s_{t}, a_{t})$. We denote the method of disjoint training as \textit{\textbf{D-READ}} in our experiments.









\subsection{Training Details}
We followed IRL for implementing text generator and reward approximator. The text generator consists of a LSTM layer with embedding size of 128 and followed by 4 linear layers with dimension of 128 along with a dropout of 0.1. We set the maximum sequence length of the generated sentences to 64. The reward approximator consists of MLPs with 3 hidden layers of 128 dimensions with a dropout of 0.2. The Classifier consists of a hidden layer of 768 dimension followed by $leaky$-$ReLu$ activation function. We have used AdamW \cite{loshchilov2018decoupled} as the optimizer with learning rate of 0.005 for $\mathcal{G}$, 0.004 for $\mathcal{R}$, and 5e-5 for both $\mathcal{M}$ and $\mathcal{C}$. 

\section{Results}
\begin{figure*}
  \centering
  \includegraphics[width=\textwidth]{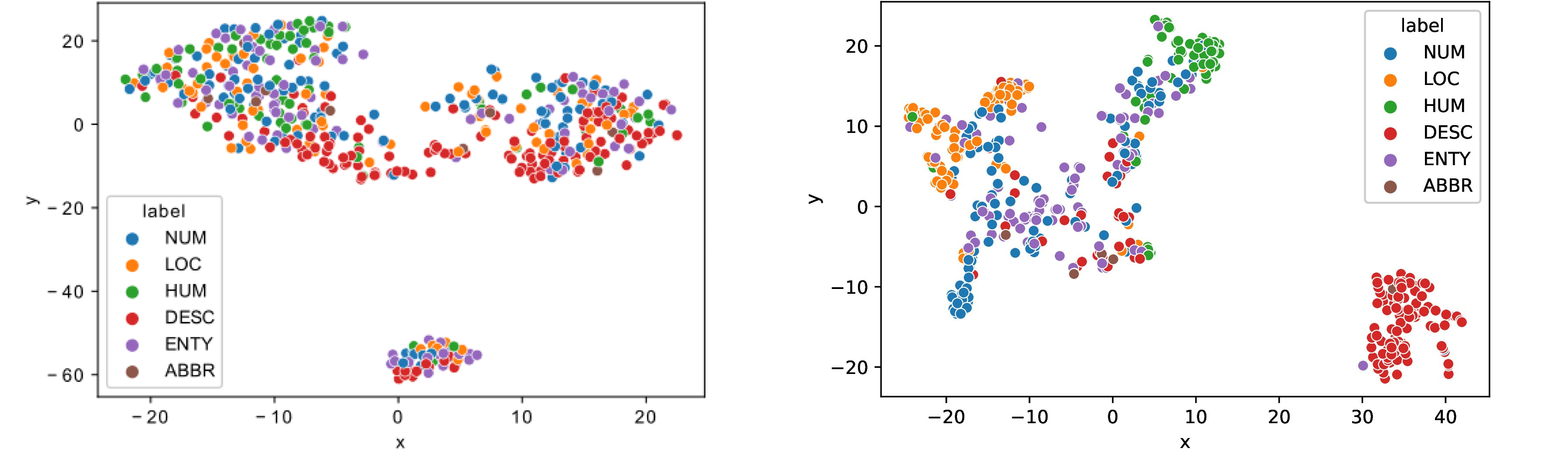}
  \caption{t-SNE Visualization of feature from using Baseline method (left) and  \textit{READ} (right) on TREC-CC task. 
  The class labels are different classes of Questions 'ABBR': Abbreviation, 'ENTY': Entity, 'DESC': Description and abstract concept, 'HUM': Human being, 'LOC': Location, 'NUM': Numeric value.
  }
  \label{fig:tsne_bert}
\end{figure*}

We have reported the accuracy for varying amount of labeled data using BERT pre-trained model in Figure \ref{fig:bert_results}. We observe that the accuracy increases with the increase in the amount of annotated data for all the methods. The Baseline method where no adversarial learning is used performs the worst among all the methods. 

On the TREC-CC task, our method \textit{READ} outperforms the GAN-BERT and Baseline method for all the values of labeled data. The gains are much more significant for the lower amount of labeled data, with gains of 68\% and 26\% over Baseline and GAN-BERT, respectively, when 2\% (108 samples) labeled data is used. Similar to GAN-BERT, the gains from our method starts to diminish with the increase in the amount of labeled data.

We observe similar trends on the SST-5 dataset with \textit{READ} outperforming all the methods in each configuration. Similar to TREC-CC task, the gains from \textit{READ} starts to diminish with the increase in the amount of annotated data. When only 1\% labeled data (85 samples) is used, our method provides 9\% and 14\% of gain over GAN-BERT and Baseline, respectively. 

TREC-CC and SST-5 datasets have only six and five classification categories. However, the TREC-CF dataset has 50 categories, making it a more challenging task than the others. The difficulty of the task is also evident from the fact that the Baseline method achieves almost 0\% accuracy when less than 2\% labeled data is used. We observe that the gains are more significant from \textit{READ} on TREC-CF than the other datasets. We also observe that the trend of diminishing gains with the increase in amount of labeled data is not visible on TREC-CF dataset, with \textit{READ} providing consistent gains for all the values of annotated data.

We have provided a similar analysis using RoBERTa in Figure \ref{fig:roberta_results} in the Appendix. We observe almost similar results to that of BERT, with slightly high accuracy in case of all the methods. It shows that irrespective of the choice of pre-trained transformer model, the proposed approach provides similar gains on all the datasets.

In Figure \ref{fig:bert_results}, we have reported the results for \textit{{D-READ}} method where the text generator and classifier are independently trained, whereas \textit{READ} encapsulates all the components through the reward function. We observe that \textit{D-READ} provides better performance than the GAN-BERT on TREC-CF dataset, showcasing the importance of text generation instead of feature generation. However, it fails to outperform \textit{READ} for all training configurations, demonstrating the importance of encapsulation of all the components.

\subsection{Generation Quality}
We hypothesize that the quality of synthetically generated text plays a big role in improving the performance of the model. To verify this, we have shown some of the generated samples by mapping it to the original text using cosine-similarity in Table \ref{table:generated_data}. The generated samples are almost similar to the real text with lot of variations, showing the diversity in the generation quality.

\subsection { Discriminative Features}
We have shown the t-SNE \cite{van2008visualizing} visualization of the features from the last layer of the BERT model after fine-tuning Baseline and \textit{READ} method on TREC-CC dataset with 1\% of labeled data in Figure \ref{fig:tsne_bert}. We can see that the features learnt from Baseline are not class-discriminative and are overlapping for lot of classes, whereas the features learnt using \textit{READ} are class-discriminative, with each cluster denoting a class-label. Our method is able to learn class-clusters with just 1\% of labeled dataset, validating the observed gains in the previous sections.


\section{Conclusion}
In this work, we propose a novel method for improving the generalization capabilities of text classifiers when fine-tuned with limited labeled data. \textit{READ} encapsulate reinforcement-based text generation and classifier through adversarial learning with the help of unlabeled data. We evaluated our method on multiple datasets and observed significant gains over the Baseline and GAN-BERT when very limited data is used. We show the importance of encapsulation through experiments and observed a significant drop in performance with disjoint training. We validated the improvements of \textit{READ} through feature visualization. Our method is only evaluated in English and can be easily extended to other languages. There have been a few works \cite{dong2019unified, li2021pretrained} proposed to improve the text generation quality by utilizing pre-trained transformer models. We plan to extend our approach by integrating these generation methods to improve performance further. 
\bibliography{anthology,custom}
\bibliographystyle{acl_natbib}

\newpage
\appendix

\section{Data Statistics}
We have provided the statistics of training and test data for each datasets in  \ref{tbl:data}. 

\begin{table}[!]
    \centering
    \small
    \setlength\tabcolsep{2pt}%
    \begin{tabular}{|p{0.3\linewidth} | p{0.2\linewidth} | p{0.15\linewidth} | p{0.15\linewidth}|}
    \hline
        \textbf{Dataset} & \#\textbf{Training Samples} & \#\textbf{Test Samples} & \#\textbf{Labels} \\ \hline\hline
        \textbf{TREC-QCF} & 5452 & 500 & 50 \\ \hline
        \textbf{TREC-QCC}  & 5452 & 500 & 6 \\ \hline
        \textbf{SST5} & 8544 & 2210 & 5 \\ \hline
    \end{tabular}
    \caption{Statistics of datasets.}
\label{tbl:data}
\end{table}

\begin{figure*}[!]
     \centering
     \begin{subfigure}[b]{0.5\textwidth}
         \centering
         \includegraphics[width=\textwidth]{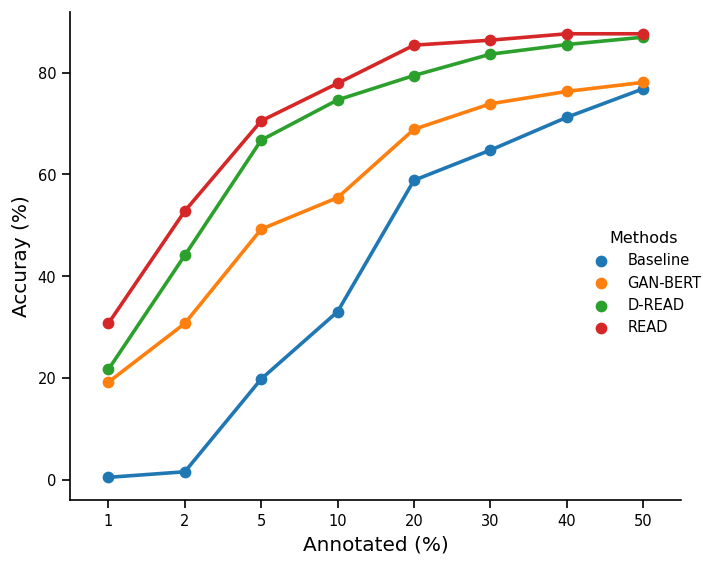}
         \caption{TREC-CF}
         \label{fig:trec_cf_roberta}
     \end{subfigure}
     \hfill
     \begin{subfigure}[b]{0.5\textwidth}
         \centering
         \includegraphics[width=\textwidth]{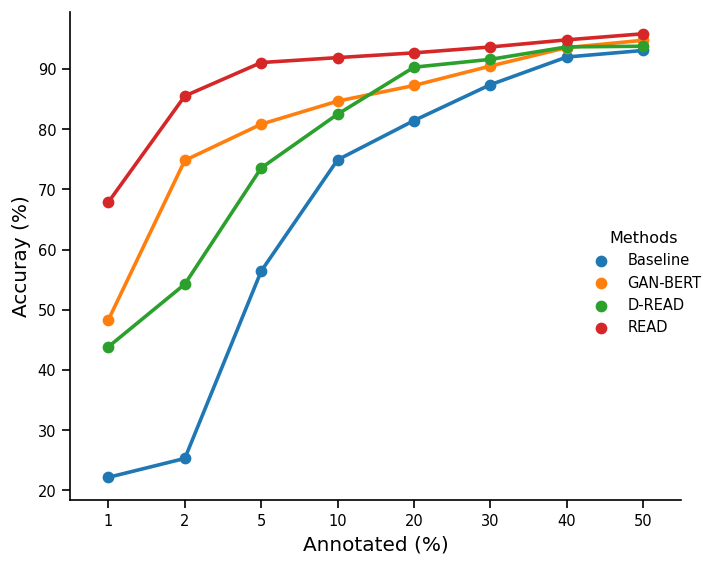}
         \caption{TREC-CC}
         \label{fig:trec_cc_roberta}
     \end{subfigure}
     \hfill
     \begin{subfigure}[b]{0.5\textwidth}
         \centering
         \includegraphics[width=\textwidth]{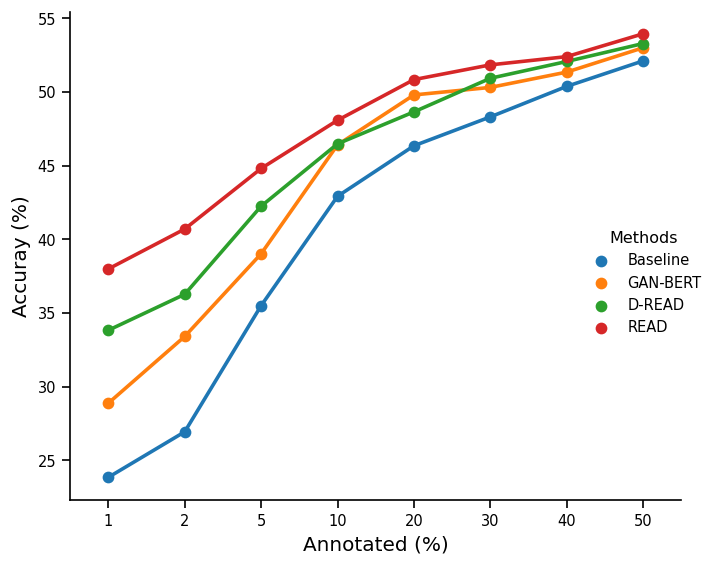}
         \caption{SST-5}
         \label{fig:sst5_roberta}
     \end{subfigure}
     \vspace{-0.5em}
      \caption{Accuracy comparison using RoBERTa model.}
        \label{fig:roberta_results}
\end{figure*}

\begin{table*}[!ht]
    \centering
    \begin{tabular}{p{0.5\linewidth}|p{0.5\linewidth}}
    \toprule
    \textbf{Real Text}  & \textbf{Generated Text} \\
    \midrule
How do they find an epicenter?	& How can I find an epicenter?\\
How do they find an epicenter?	& How can I find about an epicenter?\\
How do you find the area of a circle?	& When do you find the area of a circle?\\
How do you find the area of a circle?	& Why do we have to find the area for a circle?\\
How does it make its colors?	& What is an example of a color?\\
How does it make its colors?	& What is a origin of color?\\
How does it make its colors?	& What color were available for?\\
How is the element strontium purified?	& How long is the element strontium purified?\\
How is the element strontium purified?	& What is the element strontium purified?\\
How many Russians have landed on the moon?	& How Russians have landed on the moon?\\
How many Russians have landed on the moon?	& How many astronauts have been on the moon?\\
How many Russians have landed on the moon?	& How many people have been on the moon?\\
How many Russians have landed on the moon?	& How many Americans landed on the moon?\\
Boxing Day is celebrated on what date?	& What was the date of Boxing Day?\\
Boxing Day is celebrated on what date?	& When was Boxing Day?\\
CPR is the abbreviation for what?	& What does CPR stand for?\\
CPR is the abbreviation for what?	& What is the meaning of `` CPR ''?\\
CPR is the abbreviation for what?	& What is the definition of ` CPR ''?\\
\bottomrule
\end{tabular}
\caption{Generated data in TREC-CC Fine dataset}
\label{table:generated_data}
\end{table*}

\end{document}